\documentclass[12pt,openany,a4paper]{book}

\usepackage{xspace}
\usepackage{tabularx}

\usepackage{shortcuts}
\usepackage{enumerate}
\usepackage{amsthm,amsmath,amssymb,bm,mathtools}
\usepackage{makecell}
\usepackage{float}
\usepackage{wasysym}
\usepackage{tcolorbox}
\usepackage{url}

\usepackage{breakurl}
\usepackage{array}
\usepackage{multirow}
\usepackage{graphicx}
\usepackage{booktabs}

\usepackage{threeparttable}
\usepackage{tikz}
\usepackage{mathtools}
\usepackage{subcaption}

\newtheorem*{problem*}{Problem}

\usepackage{enumitem}

\usepackage{chngcntr}
\counterwithout{figure}{chapter}
\counterwithout{table}{chapter}

\usepackage[ruled,vlined,linesnumbered]{algorithm2e}
\SetKw{Or}{or}
\SetKw{True}{True}
\usepackage{amssymb, amsthm}
\usepackage{float}
\usepackage{wasysym}
\usepackage{tcolorbox}
\definecolor{BLUE}{rgb}{0,0,1}

\usepackage{hyperref}
\hypersetup{
  colorlinks,
  linkcolor={blue!70!green},
  citecolor={green!70!blue},
  urlcolor={orange!70!red}
}

\usepackage{filecontents}
\usepackage{marvosym}

\renewcommand{\baselinestretch}{1.2}	

\SetKwComment{Comment}{$\triangleright$\ }{}

\begin{document}

\frontmatter

\begin{titlepage}
\renewcommand{\baselinestretch}{1.0}
\begin{center}
\vspace*{35mm}
\Huge\bf
            A Slice or the Whole Pie? \\ Utility Control for AI Models\\
            \vspace{8cm}
\large\sl
		\textbf{Ye Tao}		\medskip\\

\large
		June, 2025	
\end{center}
\end{titlepage}


\chapter{Abstract}

Training deep neural networks (DNNs) has become an increasingly resource-intensive task, requiring large volumes of labeled data, substantial computational power, and considerable fine-tuning efforts to achieve optimal performance across diverse use cases. Although pre-trained models offer a useful starting point, adapting them to meet specific user needs often demands extensive customization, and infrastructure overhead. This challenge grows when a single model must support diverse applications with differing requirements for performance. Traditional solutions often involve training multiple model versions to meet varying requirements, which can be inefficient and difficult to maintain.

In order to overcome this challenge, we propose \textsc{NNObfuscator}, a novel utility control mechanism that enables AI models to dynamically modify their performance according to predefined conditions. It is different from traditional methods that need separate models for each user. Instead, \textsc{NNObfuscator} allows a single model to be adapted in real time, giving you controlled access to multiple levels of performance. This mechanism enables model owners set up tiered access, ensuring that free-tier users receive a baseline level of performance while premium users benefit from enhanced capabilities. The approach improves resource allocation, reduces unnecessary computation, and supports sustainable business models in AI deployment.

To validate our approach, we conducted experiments on multiple tasks, including image classification, semantic segmentation, and text to image generation, using well-established models such as ResNet, DeepLab, VGG16, FCN and Stable Diffusion. For image classification, we use the datasets CIFAR-10, CIFAR-100 and MNIST; for segmentation, we tested on PASCAL VOC 2012 and Cityscapes. Experimental results show that \textsc{NNObfuscator} successfully makes model more adaptable, so that a single trained model can handle a broad range of tasks without requiring a lot of changes. Our method supports scalable, cost-effective AI solutions that meet a wide range of user needs by providing an easy way to manage model performance on the fly. This is great for researchers who need to change models quickly and businesses that want to use AI to help them make decisions.

\tableofcontents

\listoffigures
\addcontentsline{toc}{chapter}{List of Figures}

\cleardoublepage

\mainmatter

\chapter{Introduction}
\label{intro}
\section{Background}
Deep neural networks (DNNs) have greatly enhanced the capabilities of artificial intelligence (AI) in fields such as computer vision, natural language processing (NLP), and medical diagnostics~\cite{karen2014very}. These models, which are often built on architectures like convolutional neural networks (CNNs) and transformers, need a lot of data and processing power to work well~\cite{lecun2015deep}.

For instance, GPT-4, which is one of the most advanced transformer-based language models, needed a lot of computing power and ran for about three months on 25,000 NVIDIA A100 GPUs. It costs about 63 million dollars to train the model once~\cite{krizhevsky2012imagenet,esteva2019guide,shen2017deep}, and it has 1.8 trillion parameters~\cite{taylor2023chatgpt}, Training ResNet-50 on the ImageNet dataset, which is a common test set in computer vision, also takes about 29 hours on a single NVIDIA V100 GPU~\cite{krizhevsky2012imagenet}. For radiology and diagnostic imaging, deep learning models in medicine often need millions of labeled medical images to work. Training can take anywhere from a few days to a few weeks, depending on the size of the dataset and the amount of computing power available~\cite{esteva2019guide,shen2017deep,wang2024core}.

Even with these improvements, it is still very hard to use and adapt AI models for different purposes~\cite{srinivas2017training}. Fine-tuning, domain adaptation, and inference optimization are examples of tasks that add a lot of complexity~\cite{sun2023shadownet}, which makes it hard to use a single pre-trained model for different tasks and computing environments~\cite{ruder2019neural}. As AI adoption grows, there is a growing demand for more scalable, efficient, and adaptable AI solutions that can reduce computational costs while maintaining strong performance across different domains~\cite{yan2024exploring,cai2019once,zoph2018learning,yan2024quality,yan2025tracking}.
A well-trained AI model is often expected to serve a wide range of users, each with different requirements regarding accuracy, computational efficiency, and functionality~\cite{tan2018survey}. However, traditional deep learning workflows typically require retraining or fine-tuning separate models to meet different user needs, significantly increasing computational costs and development time~\cite{tan2018survey,wang2025aim,wang2023data}. Furthermore, maintaining multiple fine-tuned models introduces additional storage and infrastructure overhead, leading to excessive resource consumption, particularly in large-scale deployments.

\section{Requirements}
As AI systems become increasingly integrated into real-world applications, two practical needs have emerged as critical factors in the deployment and sustainability of these technologies: personalized adaptation for end-users and strategic tiered access for service providers. Meeting these requirements not only improves user satisfaction, but also helps make AI systems more flexible, fair, and easier to use in different situations.

\textbf{Personalized Adaptation for Users}. People who use the same AI system may need different levels of model performance. Some people may need very high accuracy, while others may be okay with lower performance as long as it meets their basic needs. This makes it hard for one model to meet all of the needs at once. For instance, in a self-driving system, one user might only need basic object detection, while another might need to be able to find pedestrians and traffic signs with great accuracy. In the same way, a city planner might want a remote sensing image analysis system to focus on finding buildings and roads to help with urban infrastructure planning, while an environmental researcher might be fine with less detailed results.

There are more examples in the healthcare field. A doctor in a hospital might have to look at a lot of medical pictures before making a choice. A user of a health app might only want to know if everything is ``normal'' or if they should ``check with a doctor.'' When it comes to educational tools, beginners might need answers that are easier and slower, while advanced learners might want answers that are more accurate and complete.

AI models should be able to change how they work to fit the needs of each person. You can change the thresholds for making predictions, use user profiles, or process users at different levels depending on what type they are.

\textbf{Strategic Model Tiering for Owners}. Many AI-powered services use a tiered access or freemium model. This means that regular users can get a basic version of an AI model that works and is accurate enough, but premium users can get access to a more accurate version. A good example is automated language translation tools. A free version might give you basic translations that are good for everyday use. On the other hand, an upgraded version might help you with terms that are specific to your field and make sure that the translations are more accurate in context with commercial or legal documents. In the same way, a simple AI model could do basic object detection in computer vision apps. While a more advanced model could do real-time tracking and predictive analytics for security surveillance. This organized method lets AI companies give away free or cheap models to get more people to use them, and then charge professionals for extra features. This makes it easier to get AI and build it.

This way of setting up models makes it easier for users to find the right one on certain platforms. It also enables developers change and update models in stages, so some users get new features while others keep using the same version. This tiered structure also makes it easier to test out new ideas, get feedback from different groups of users, and make sure that more people can use AI without always needing the most powerful tools.

In real-world applications, not all users want the same thing from AI models. Some people care more about speed than others, while others focus on detailed and accurate results, but might only need a simple answer. At the same time, building flexible models with different output lets AI to help more people. These two needs personalized output and multiple model levels are important for making AI useful, practical, and easier to manage in real life.

Transfer learning, knowledge distillation, dynamic model pruning, and multi-task learning~\cite{hinton2015distilling,he2020learning,gou2021knowledge} are just a few of the new AI optimization techniques that have come out recently to make models more efficient and adaptable. These methods have improved AI performance in a number of ways, they remain insufficient in addressing two important requirements: the ability of a single model to simultaneously meet different user needs with dynamic performance, and the ability for model owners to offer tiered AI services without requiring extensive retraining or multiple model deployments~\cite{wang2025aim}.

These techniques make things better, but they don't fully solve the main problem of dynamically controlling AI models based on changing user needs and business plans. Static training pipelines are a big part of how AI is used today. Current AI deployment strategies rely heavily on static training pipelines, this means that to adapt a model to different tasks or user preferences you often have to get separate models back, retraining them, or redeployment of separate models. This not only makes AI systems less flexible, but it also adds a lot of costs to running and computing~\cite{wang2025aim}. Models will keep having trouble finding the right balance between computational efficiency, user-specific task prioritization, and smooth business growth until AI frameworks become more flexible. Future AI improvements should focus on making it possible to change models in real time, adapt tasks to each user, and unlock features on the fly, all without having to retrain a lot of models or use a lot of different models.

\section{Our Work}
The \textsc{NNObfuscator}: Utility Control for AI Models project offers a new way to deal with two common problems that come up when deploying AI models: (1) allowing a single model to meet the needs of different users by performing tasks in different ways, and (2) allowing model owners to offer tiered AI services without having to retrain the model or deploy multiple models. Unlike traditional approaches such as transfer learning, knowledge distillation, and model pruning, which require significant modifications and retraining, \textsc{NNObfuscator} applies controlled perturbations to pre-trained neural networks to adjust their utility dynamically.

\begin{figure}[h!]
    \centering
        \includegraphics[width=\linewidth]{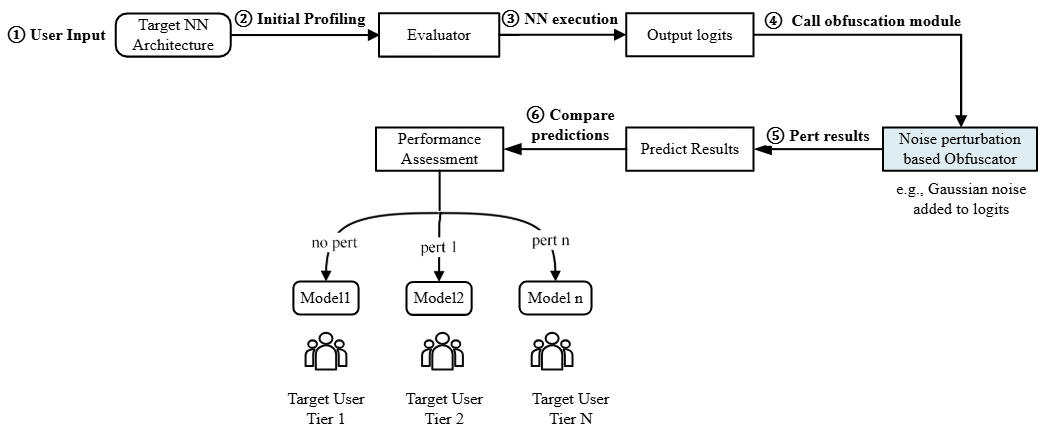}
    \caption{Noise Perturbation Based Processing Workflow in Neural Network Models}
    \label{fig:fig1}
\end{figure}

\add{}

As shown in Fig.~\ref{fig:fig1}, this method enables a flexible mechanism for AI models to adapt their behavior based on user preferences in real time, without compromising their core structure. For example, \textsc{NNObfuscator} could help a medical AI system give radiologists a model that focuses on finding cancer with high accuracy, while also giving general practitioners a model that is easier to understand and less complex. Similarly, in a financial fraud detection system, a bank may deploy a model that prioritizes high precision for high-value transactions while using a more efficient, lower-latency version for real-time small transaction monitoring.  

\begin{figure}[h!]
    \centering
        \includegraphics[width=0.6\linewidth]{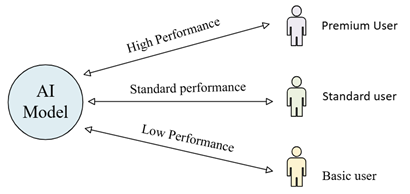}
    \caption{Secnario \#1: Single model to dynamic task performance}
    \label{fig:figure2}
\end{figure}

This addresses the first key demand by offering fine-grained control over model utility, ensuring different users can obtain tailored performance without requiring separate model versions, as shown in Fig.~\ref{fig:figure2}.
 
\begin{figure}[h!]
    \centering
        \includegraphics[width=0.6\linewidth]{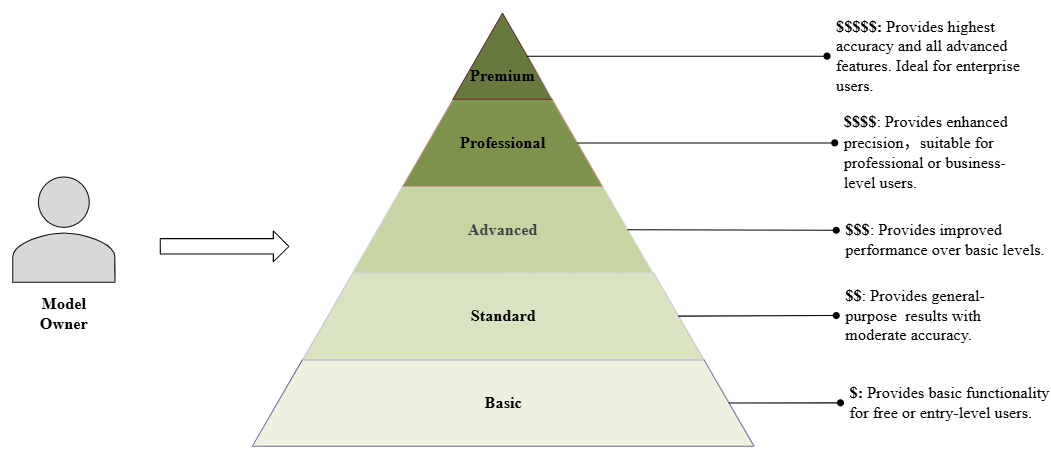}
    \caption{Secnario \#2: Tiered AI services}
    \label{fig:figure3}
\end{figure}

From the perspective of model owners, \textsc{NNObfuscator} also facilitates a tiered AI service model, where varying levels of model performance and capabilities can be offered dynamically. Instead of developing distinct models for free, standard, and premium users, a single core model can be adjusted in real time to unlock advanced features or higher accuracy for paying customers. This is particularly valuable in AI-driven SaaS platforms, where model monetization and scalability are critical.

\section{Challenges}
Even though it has some benefits, the \textsc{NNObfuscator} method comes with a number of technical problems that need to be carefully solved to make sure it works and is easy to use. One of the biggest problems is how to intentionally lower the performance of a model in a controlled way. The goal is to change only the predictions that were right at first, which will lower the overall accuracy of the predictions, while making sure that predictions that were wrong before stay the same. This calls for a smart perturbation strategy that focuses on high-confidence correct predictions instead of adding random noise to the whole model. Obfuscation can make things worse for no reason if you don't know what you're doing, which is not good or efficient. \textsc{NNObfuscator} fixes this by using a precision-targeted perturbation mechanism that changes decision boundaries in a consistent way for predictions that are very likely to be correct. This makes sure that the model's accuracy goes down in a way that can be predicted and controlled. This gives you control over how useful the model is without making the results less stable.

Another big problem is making sure that changes only affect important parts of an input and don't randomly break the whole model. A lot of AI models, especially those that deal with computer vision and natural language processing, treat inputs differently depending on where they are. When you sort pictures or break them up into parts based on what they mean, the most important things in the picture are the people or cars in the picture, not the things in the background. A simple obfuscation method that affects all areas equally would lower accuracy without taking into account how important the context is, which could make things harder to use and cause problems with accuracy. 

\textsc{NNObfuscator} fixes this by using a perturbation strategy that knows about regions. This means it changes important areas but leaves less important or background areas alone. This means that accuracy degradation is only used when it matters most, not to make the whole input louder. In the real world, this method lets model owners choose how much worse some important predictions get. This makes sure that obfuscation meets business and security goals while causing as few problems as possible.

Our work presents a new way to control how AI models use utility at a very small level. It fixes a lot of the problems that older methods like transfer learning, knowledge distillation, and model pruning had. You have to do a lot of retraining, fine-tuning, or making multiple static models with these methods. \textsc{NNObfuscator}, on the other hand, adds a dynamic and controlled obfuscation mechanism that lets a single model change how useful it is depending on what different users need. This new idea meets two important needs right away: (1) it lets a single AI model meet the needs of different users by changing the order in which tasks are done, and (2) it lets model owners use a tiered AI service model without having to do a lot of retraining or separate deployments.

\section{Contributions}
One of \textsc{NNObfuscator}'s main features is that it can intentionally make a model work worse. This means that the model can be made more useful by changing predictions that were right at first, but predictions that were wrong at first stay the same. This is a big step up from old methods, where model degradation is often random or out of control, which makes it hard to predict behavior and bad for users. A precision-targeted perturbation mechanism makes sure that obfuscation doesn't make things any more unstable while still letting users change the accuracy of the system to fit their needs.

This work also makes a regional aware perturbation strategy that only changes areas that are very important while leaving areas that are not as important the same. This is very important for both computer vision and natural language processing because random obfuscation could make models very hard to work with. \textsc{NNObfuscator} is a great way to control how well an AI model works without changing its basic structure or putting more work on the computer. It does this by only using perturbations where they matter most.

\textsc{NNObfuscator} has new technical features, and it also gives you new ways to make money with AI models and use them. It lets AI companies use business models that can grow, like free-tier models with limited accuracy, premium versions with better performance, or customizable models that let users change how their AI works. This is because it lets them change how well the model works right away. This is a big improvement over how AI is usually used, which requires training, maintaining, and sharing different models for each level of service. \textsc{NNObfuscator} lets AI companies offer personalized AI solutions on a large scale. This makes the solutions more affordable and easier to get to, while also making them more flexible.

This study also adds to the larger field of AI security and model control because obfuscation techniques can also be used to protect proprietary AI models from attacks and unauthorized use. \textsc{NNObfuscator} is a two-in-one framework that makes both customization and model protection better by combining utility control with security-boosting methods. Because of this, it is very useful for AI in healthcare, finance, self-driving cars, and other fields.

\section{Objectives \& Significance}
The main goal of this study is to make \textsc{NNObfuscator}, a new way to use targeted perturbation techniques to control how useful AI models are. \textsc{NNObfuscator} is not the same as old methods that use static fine-tuning, retraining, or deploying more than one model version. It gives you a flexible and scalable solution that lets a single AI model change in real time to fit the needs of different users and business plans. 
The main things this study wants to do are:

\begin{itemize}
\item Implement a region-aware obfuscation strategy. Selectively apply perturbations to high-significance regions while keeping background or low-priority areas intact, preserving essential model functionality while allowing flexible accuracy control.

\item Enable dynamic user-controlled AI adaptation. Design a framework where users or model owners can adjust model performance on demand, allowing different levels of accuracy, interpretability, or efficiency depending on real-time needs.

\item Support a scalable AI business model for model owners. Provide AI service providers with a tiered deployment strategy, allowing the same model to be offered at different accuracy levels (e.g., free-tier vs. premium) without requiring multiple separate models or extensive retraining.
\item Ensure robustness and security of the obfuscation approach. Evaluate the impact of obfuscation on model security and resistance to adversarial attacks, ensuring that perturbation does not introduce vulnerabilities that could be exploited by malicious actors.
\end{itemize}

\chapter{Related Work}
The development of \textsc{NNObfuscator} is related to various techniques in deep learning, particularly those focused on model adaptation, knowledge transfer, and representation learning. These techniques aim to modify or improve neural networks by adjusting their learned representations while maintaining or enhancing their performance. While \textsc{NNObfuscator} shares similarities with these approaches, it differs in its focus on modifying model behavior through targeted obfuscation rather than improving generalization or transferability.

 \begin{figure}[h!]
    \centering
        \includegraphics[width=0.5\linewidth]{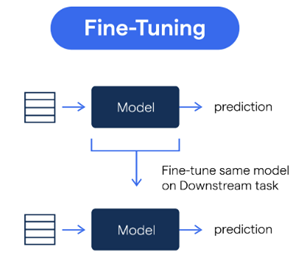}
    \caption{Fine-tuning processing~\cite{RN113}.}
    \label{fig:figure4}
\end{figure}

\section{Fine-Tuning}
One closely related area is Fine-Tuning, a technique widely used in deep learning where a pre-trained model is adapted to a new task by updating only part of its weights or retraining the entire model with a smaller learning rate~\cite{liu2015deep,howard2018universal}. Fine-tuning helps models retain prior knowledge while adapting to new data~\cite{kornblith2019better,thimm1995evaluating}, making it a powerful tool in scenarios with limited training samples~\cite{you2019large,hu2022lora,zhang2021survey}. However, \textsc{NNObfuscator} differs by intentionally obfuscating specific learned representations rather than refining them, allowing for controlled degradation of model performance to study feature dependencies~\cite{liu2018fine}.

 \begin{figure}[h!]
    \centering
        \includegraphics[width=0.7\linewidth]{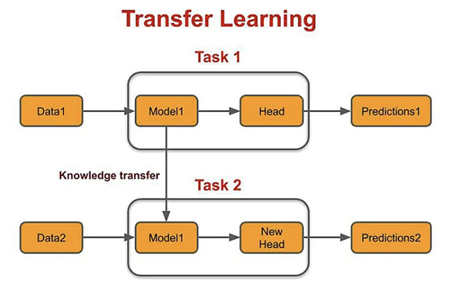}
    \caption{A diagram describing Transfer Learning~\cite{RN114}.}
    \label{fig:figure5}
\end{figure}

\section{Transfer Learning}
Another relevant technique is Transfer Learning, which enables models trained on one domain to be applied to another by leveraging shared feature representations~\cite{long2015learning,torrey2010transfer}. Methods such as domain adaptation and feature extraction help improve performance in new tasks by reusing knowledge from a source domain~\cite{long2015learning,torrey2010transfer}. While transfer learning focuses on preserving useful features, \textsc{NNObfuscator} instead disrupts critical features to analyze their impact on the model's performance, making it a tool for robustness analysis rather than knowledge reuse~\cite{yosinski2014transferable}.

\section{Feature Distillation}
A third related approach is Feature Distillation, commonly used in knowledge distillation and model compression, where a smaller model learns to mimic the behavior of a larger model by extracting and utilizing essential features~\cite{sun2019patient,wang2025feature}. Feature distillation aims to maintain the most important information while reducing redundancy in representations~\cite{gou2021knowledge}. \textsc{NNObfuscator}, on the other hand, works in the opposite direction by intentionally masking or altering specific feature representations to evaluate how much a model depends on them.

\section{Multi-task learning}
Multi-task learning (MTL) trains a model to handle multiple tasks simultaneously~\cite{standley2020tasks}, which can improve efficiency when related tasks share underlying features~\cite{guo2018dynamic}. However, this method lacks real-time task prioritization capabilities, meaning that while a model may be capable of handling multiple objectives, it cannot dynamically adjust its focus based on user input~\cite{tao2023task}. For instance, an AI used in autonomous Vehicles can be taught to recognize both road signs and people walking. Even though it can technically do both tasks, a user can't easily tell the system to switch priorities between the two in real time without a lot of reconfiguration~\cite{crawshaw2020multi}.
 

 \begin{figure}[h!]
    \centering
        \includegraphics[width=0.5\linewidth]{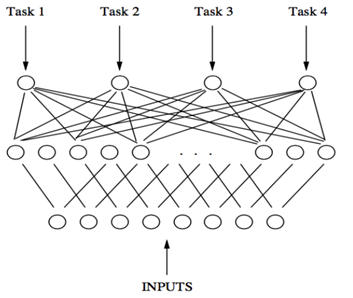}
    \caption{Applications for Multitask Learning~\cite{RN115}.}
    \label{fig:figure6}
\end{figure}

\section{Network-level Obfuscation}
Obfuscation neural networks are a new way to make deep learning (DL) models safer and easier to work with in the last few years. In order to use this method, you have to change or disturb an existing DL model in some way. This makes it harder to figure out or copy the model's internal structure or parameters, but it doesn't change how the model works or how well it works~\cite{chen2018protect}. Obfuscation makes the model's internal details less clear, which makes it safer and easier to control~\cite{papernot2018sok,shokri2017membership,liu2024purpose}.

\noindent Advantages:
\begin{itemize}
\item Enhanced Security: Obfuscation adds an extra layer of protection against unauthorized use or reverse engineering by making it harder to understand or copy how the model works inside~\cite{abadi2016deep,tramer2016stealing}.
\item Control: You can change how much obfuscation there is, which lets you customize the model for different users without losing its usefulness as a whole~\cite{fredrikson2015model}.
\item Customizability: You can change how much obfuscation there is, which lets you customize the model for different users without losing its usefulness as a whole.
Disadvantages:
\item Possible Loss of Transparency: When things are hidden, they can be harder to understand. This can be a problem in situations where you need to understand how a model works.
\item Trade-offs in performance: The model's basic functions will still work, but the extra complexity of obfuscation may make it less accurate or require more processing power.   
\end{itemize}


 \begin{figure}[h!]
    \centering
        \includegraphics[width=\linewidth]{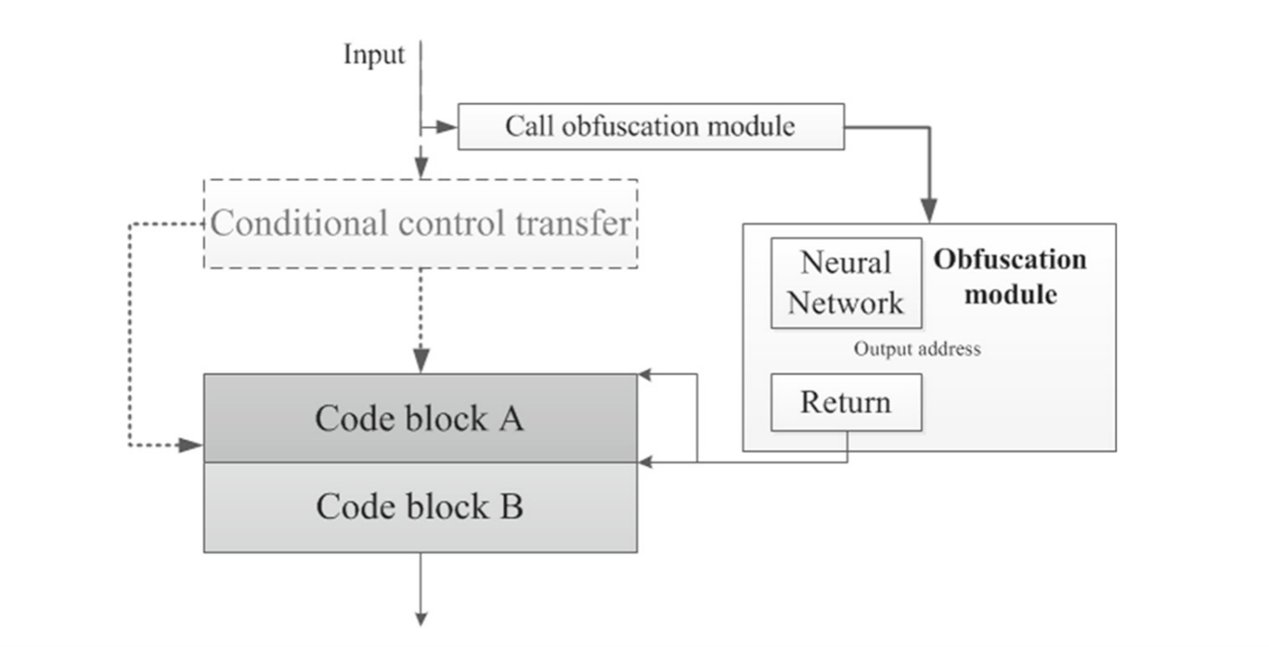}
    \caption{Semantic obfuscation using neural network~\cite{chen2018semantic}.}
    \label{fig:figure7}
\end{figure}

\section{Model level Obfuscation}
Researchers have been working on new ways to make DL models more adaptable and customizable without changing the basic structure of the models. This is because previous DL models had some problems. One way to do this is to add obfuscation techniques to deep learning models that have already been trained. Obfuscation means adding controlled changes to a model's output. This lets you make different versions of the model that meet the needs of different users while keeping the original model's performance and structure~\cite{hendrycks2019benchmarking,wang2025aim,zhongkui2023formal}.

There are many benefits to using the newest DL models with obfuscation. First, they let you control how the model is used at the model level, so you can change how useful it is for each user without having to retrain it from scratch. This method not only saves time and computer power, but it also makes the model more flexible in different situations~\cite{he2016deep,carlini2017towards}.

There are, however, some problems with using obfuscation in DL models. One of the main problems is making sure that the changes made don't make the model's performance worse or cause problems like biased outputs. Also, we need strong ways to test how well obfuscation techniques work in different situations, especially in real-world applications where model accuracy and reliability are very important~\cite{yan2025understanding,yan2024investigating}.

\chapter{Problem Formulation}
\section{Deep Neural Networks (DNNs)}
DNNs have become the foundation of modern artificial intelligence, demonstrating remarkable capabilities across a wide range of applications, including computer vision, natural language processing, healthcare, finance, and autonomous systems~\cite{devlin2019bert,litjens2017survey}. A typical DNN consists of multiple layers of neurons, each performing nonlinear transformations on input data, enabling hierarchical feature extraction. Mathematically, a DNN can be represented as a function:
\begin{align*}
z^{(1)} & =\sigma\left(W^{(1)} x+b^{(1)}\right) \\
z^{(2)} & =\sigma\left(W^{(2)} z^{(1)}+b^{(2)}\right) \\
\vdots & \\
z^{(L)} & =\sigma\left(W^{(L)} z^{(L-1)}+b^{(L)}\right) \\
y & =z^{(L)}
\end{align*}
$W^{(l)}$ is the weight matrix of the $l$-th layer, $b^{(l)}$ is the bias vector, $z^{(l)}$ is the output of layer, $\sigma(\cdot)$ is a nonlinear activation function like ReLU or sigmoid, $y$ is the final output of the network. The network learns the weights and biases during training by minimizing the difference between the predicted output and the true output on a training dataset. This is usually done using a method called gradient descent.
 

 \begin{figure}[h!]
    \centering
        \includegraphics[width=\linewidth]{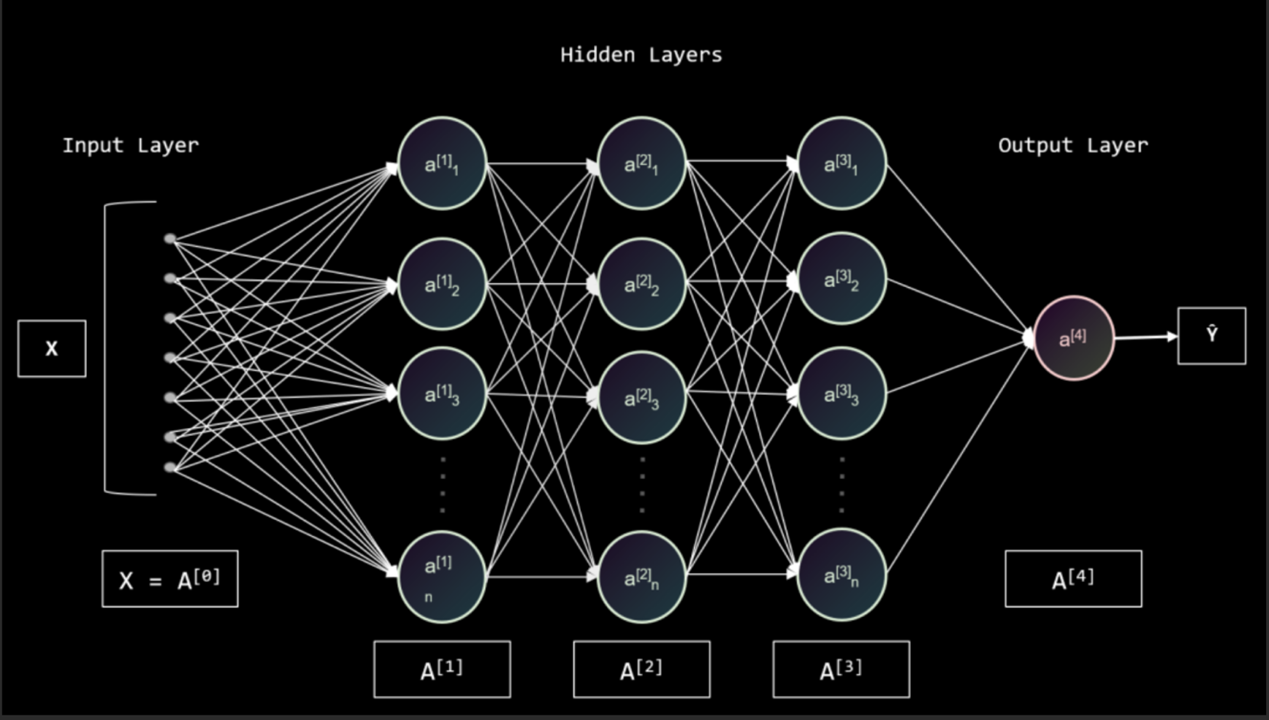}
    \caption{An example of neural network~\cite{gusak2022survey}.}
    \label{fig:figure8}
\end{figure}

Despite their effectiveness, DNNs are resource-intensive and often trained for a single optimized purpose, making them rigid in dynamically adjusting to diverse user needs~\cite{cai2019once}. The challenge lies in adapting a single trained model for different users or business strategies without requiring multiple versions or extensive retraining~\cite{laskaridis2021adaptive}.

\section{Motivation}
As AI adoption increases, different users have varying requirements for model utility, creating a fundamental challenge in AI deployment. In real-world scenarios:
\begin{itemize}
\item Users demand different levels of performance based on their specific needs. For example, a doctor using a medical AI system may require high accuracy in detecting life-threatening diseases, while a hospital administrator may prefer faster but less precise predictions for large-scale patient screenings.
\item AI service providers need a scalable business model, where a single AI model can be offered at different performance tiers (e.g., basic, advanced, premium) without incurring additional computational and storage costs from maintaining multiple trained versions.
\end{itemize}
Traditional methods, such as transfer learning, fine-tuning, and knowledge distillation as mentioned before, do not address this challenge effectively because they require:
\begin{itemize}
\item Extensive retraining, which is computationally expensive.
\item Deployment of multiple separate models, leading to inefficient resource utilization.
\item Lack of real-time control, as model modifications are static and not user-adjustable.
\end{itemize}

\section{Defining \textsc{NNObfuscator}}
\textsc{NNObfuscator} is a utility control framework for DNNs that applies targeted perturbations to regulate model performance without modifying the core model parameters or requiring retraining. The approach allows adaptive accuracy control, enabling model owners to customize the model's performance in real-time. The core objectives of \textsc{NNObfuscator} are:
\begin{enumerate}
\item Targeted Prediction Degradation
\begin{itemize}
\item Instead of randomly introducing errors, \textsc{NNObfuscator} intelligently modifies only originally correct predictions to reduce accuracy while keeping originally incorrect predictions unchanged.
\item This is achieved through a prediction-sensitive perturbation function:
\begin{align*}
\tilde{y}=P(y, \delta)= \begin{cases}y^{\prime}, & \text { if } y=\hat{y} \text { (originally correct prediction) } \\ y, & \text { otherwise }\end{cases}
\end{align*}
\end{itemize}
where:
$y$ is the original prediction, $\hat{y}$ is the ground truth label,
$P(y,\delta)$ is the perturbation function with obfuscation strength, $\tilde{y}$ is the modified prediction, $y^\prime$ is the newly perturbed incorrect prediction.
This mechanism makes sure that prediction accuracy is controlled in a systematic way rather than degrading randomly, which lets for precise utility adjustments.
\item Region-Aware Perturbation Strategy
\end{enumerate}

\begin{itemize}

\item In applications such as image classification or semantic segmentation, not all parts of an input contribute equally to a decision. \textsc{NNObfuscator} selectively applies perturbations to high-significance regions while leaving background areas intact.
\item The perturbation mask M(x) is computed based on importance scores I(x):
\begin{align*}
M(x)= \begin{cases}1, & \text { if } I(x)>\tau \\ 0, & \text { otherwise }\end{cases}
\end{align*}
\end{itemize}
\noindent where:
$I(x)$ is an importance function that measures the contribution of input $x$ to the final decision,
$\tau$ is a threshold determining the significance of each region,
$M(x)$ controls where perturbations are applied.

\section{Adaptive Model Utility Control}
\textsc{NNObfuscator} adds Gaussian noise to the model's predictions to give you dynamic control over how useful the model is. This changes the accuracy levels in a smooth and controllable way.
The modified prediction ($x,\theta,\delta$) is given by:
\begin{align*}
\tilde{f}(x, \theta, \delta)=f(x, \theta)+\delta \cdot \mathcal{N}\left(0, \sigma^2\right)
\end{align*}
\noindent where:
$f(x, \theta)$ is the original model prediction,
$\mathcal{N}(0, \sigma^2)$ represents Gaussian noise with mean 0 and variance $\delta$ is the obfuscation strength, controlling the amount of noise added to the prediction.
\textsc{NNObfuscator} adds Gaussian noise to the model's output to make sure that it keeps getting worse and can be controlled, so users can choose how useful they want it to be. If you set $\delta$ to a higher value, the accuracy will go down more. If you set it to a lower value, the performance will stay closer to the original model.

\chapter{Methodology}
To better manage the trade-off between model performance and data privacy, we suggest a structured method that uses noise perturbation techniques. The \textsc{NNObfuscator} framework is meant to add controlled noise to the logits of a neural network, thereby achieving a tunable trade-off between accuracy (for classification tasks) and mean Intersection over Union (mIoU, for segmentation tasks). The approach consists of three main components:
\begin{itemize}
\item \textbf{Logits Perturbation (use Gaussian noise)}: Adding independent Gaussian noise to each logit output to degrade accuracy in a controlled manner.
\item \textbf{Region-Focused Logits Perturbation (use Grad CAM)}: Applying noise selectively to important regions of an image using enhancement maps, rather than uniformly modifying all logits.
\item \textbf{Mapping via a Fitted Function}: Learning an empirical relationship between the noise level and accuracy/mIoU to enable precise noise adjustment for a given target accuracy or segmentation performance.
\end{itemize}
By combining these three aspects, \textsc{NNObfuscator} provides a flexible mechanism for obfuscating neural network predictions while maintaining the ability to tune performance degradation to a desired level.

\section{Logits Perturbation}
 

 \begin{figure}[h!]
    \centering
        \includegraphics[width=\linewidth]{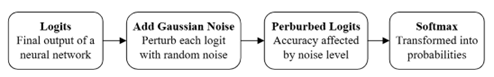}
    \caption{Logits perturbation.}
    \label{fig:figure9}
\end{figure}

For standard classification tasks, the final output of a neural network is typically a vector of logits, which represent the unnormalized class scores, denoted as $z=[z_1,z_2,…,z_K]$, where each $z_i$ represents the unnormalized score for class $i$, and $K$ is the number of classes. These logits are then transformed into probabilities using the softmax function. The softmax function ensures that each output probability is in the range (0,1), the sum of all output probabilities is 1. The softmax transformation is defined as~\cite{lu2021soft}: 
\begin{align*}
p_i=\frac{\exp \left(z_i\right)}{\sum_{j=1}^K \exp \left(z_j\right)}, \quad \text { for } i=1,2, \ldots, K
\end{align*}
or in vector form~\cite{zhu2020efficient}:
\begin{align*}
\mathbf{p}=\operatorname{softmax}(\mathbf{z})=\left[\frac{e^{z_1}}{\sum_{j=1}^K e^{z_j}}, \frac{e^{z_2}}{\sum_{j=1}^K e^{z_j}}, \ldots, \frac{e^{z_K}}{\sum_{j=1}^K e^{z_j}}\right]
\end{align*}
where $z_i$ is the logit corresponding to class $i$. $exp(z_i)$ is the exponential of the $i$-th logit,  $\sum_{j=1}^K e^{z_j}$ normalizes the logits across all classes.
To introduce controlled obfuscation, we perturb each logit with independent Gaussian noise~\cite{luisier2010image}:
\begin{align*}
\tilde{z}_i=z_i+\mathcal{N}\left(0, \sigma^2\right)
\end{align*}
where $\mathcal{N}\left(0, \sigma^2\right)$ represents a normal distribution with zero mean and variance $\sigma^2$. The parameter $\sigma$ controls the level of randomness introduced to the logits, thereby affecting the accuracy of the model's predictions.

\subsection{Effect of Gaussian Noise on Accuracy}
Introducing noise at the logits level impacts the classification performance in a predictable way. As $\sigma$ increases, the perturbation can cause:
\begin{itemize}
\item Misclassification of previously confident predictions.
\item Increased uncertainty in decision boundaries.
\item Reduction in overall accuracy.
\end{itemize}
Through empirical evaluations, we fit a function $A=f(\sigma)$ that characterizes the relationship between the noise level $\sigma$ and the resulting accuracy $A$. This allows us to select an appropriate noise level to achieve a desired classification accuracy.

\subsection{Adaptive Noise Injection for Controlled Accuracy Degradation}
Given a target accuracy $A_{target}$, we solve for the required noise level:
\begin{align*}
\sigma=f^{-1}(A_{target})
\end{align*}
This function allows \textsc{NNObfuscator} to dynamically adjust the noise level to achieve the desired accuracy degradation.

\section{Region-Focused Logits Perturbation}
Unlike classification tasks, semantic segmentation requires pixel-wise predictions, meaning that indiscriminate perturbation of all logits could lead to undesirable artifacts in the output. Instead, we propose a region-focused noise injection strategy based on enhancement maps.
To selectively apply noise, we generate an enhancement map M that highlights important regions of an image. This can be derived from attention-based methods (e.g., Grad-CAM)~\cite{selvaraju2017grad,chen2020adapting}.
 

 \begin{figure}[h!]
    \centering
        \includegraphics[width=0.8\linewidth]{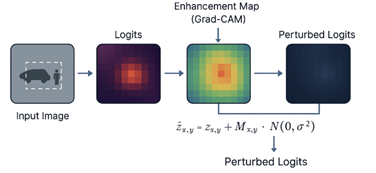}
    \caption{Region focused logits perturbation.}
    \label{fig:figure10}
\end{figure}

Using the enhancement map, we apply noise selectively:
\begin{align*}
\tilde{z}_{x, y}=z_{x, y}+M_{x, y} \cdot \mathcal{N}\left(0, \sigma^2\right)
\end{align*}
where $M$ modulates the noise level based on region importance. This ensures: key regions receive moderate noise, preserving their segmentation quality. Less important regions receive stronger perturbations, contributing to obfuscation. 
 

 \begin{figure}[h!]
    \centering
        \includegraphics[width=0.5\linewidth]{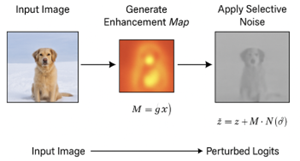}
    \caption{Example for region-focused logit perturbation.}
    \label{fig:figure11}
\end{figure}

\section{Mapping via a Fitted Function}
To provide a systematic way of tuning noise levels for both classification and segmentation tasks, we empirically determine the relationship between noise and accuracy/mIoU.

\subsection{Function Fitting for Accuracy Degradation}
For classification, we conduct a series of experiments to establish an empirical relationship $A=f(\sigma)$,where $A$ is the accuracy and $\sigma$ is the noise level.
Using regression techniques, we fit a function $f(\sigma)$ that allows us to determine the required noise level $\sigma$ for a given target accuracy: $A_{target:\;\sigma=f^{-1}(A_{target})}$

\subsection{Function Fitting for Segmentation Quality (mIoU)}
Similarly, for semantic segmentation, we establish a function that maps noise level $\sigma$ to the mean Intersection-over-Union (mIoU): mIoU=$g(\sigma)$.
By inverting this function, we can determine the required noise level to achieve a target mIoU: $miou(\sigma)=a\cdot e^{-b\sigma}+c$, $a$ determines the scale of the decay, $b$ controls the rate of degradation, $c$ is the performance floor (the lowest mIoU under heavy noise). The parameters are learned via nonlinear curve fitting using the fit function, as shown below:

 \begin{figure}[h!]
    \centering
        \includegraphics[width=\linewidth]{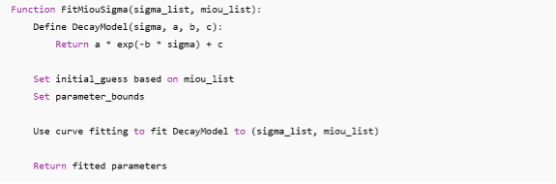}
    \label{fig:figpp}
\end{figure}

\chapter{Experimental Evaluation}
We ran a lot of tests on different deep learning architectures and datasets to see how well \textsc{NNObfuscator} worked. There are two main parts to our evaluation: classification tasks, where we look at how accurate the results are, and semantic segmentation tasks, where we look at how noise affects mIoU. The datasets and models that were chosen make sure that there are a lot of different testing situations, such as different kinds of images and levels of model complexity.
6.1 Datasets and Models
We used a mix of well-known datasets and deep learning architectures to see how \textsc{NNObfuscator} affected different tasks.
 \begin{figure}[h!]
    \centering
        \includegraphics[width=0.8\linewidth]{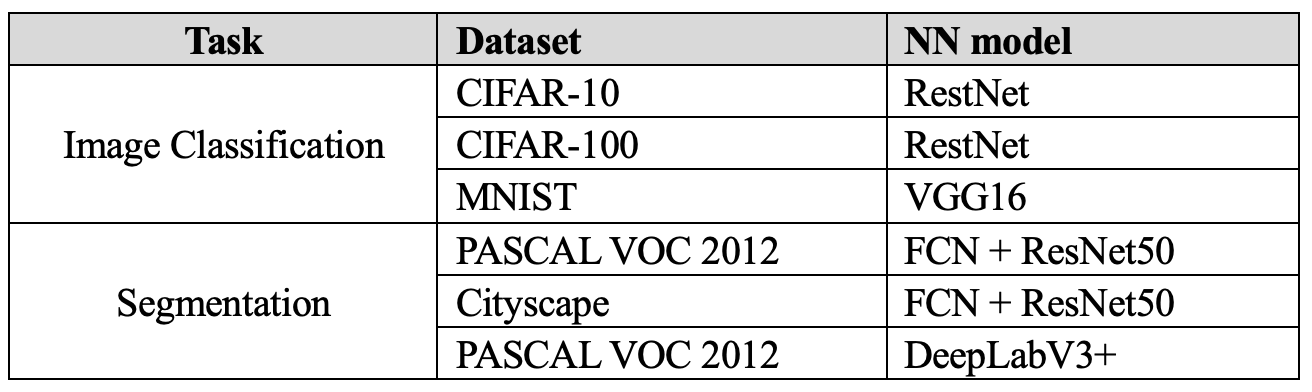}
    \label{fig:figtab}
\end{figure}
The datasets have tasks that are easy, like recognizing digits, and tasks that are harder, like separating objects and scenes. These tasks show how hard computer vision can be at different levels. The models that were chosen are well-known and do a good job of separating and classifying things. This is why they are great for testing \textsc{NNObfuscator}'s ability to change how well a model works on the fly.

\paragraph{Classification Datasets}
\textbf{CIFAR-10}: A dataset with 60,000 images in 10 classes, such as airplanes, birds, cats, and trucks, that is used a lot. The dataset is hard for deep models because each image is only 32×32 pixels~\cite{krizhevsky2009learning}.

Model Used: \textbf{ResNet} (Residual Networks), a deep convolutional neural network (CNN) known for its skip connections, which help with gradient flow during training~\cite{he2016deep}.

\textbf{CIFAR-100}: Like CIFAR-10 but with 100 classes, making it significantly more challenging. Each class has fewer samples, leading to increased model complexity~\cite{krizhevsky2009learning}.

Model Used: \textbf{ResNet}, as it is well-suited for deep learning tasks with large numbers of categories~\cite{szegedy2016rethinking}.

\textbf{MNIST}: A simple yet effective dataset of handwritten digits (0-9), containing 28×28 grayscale images. Frequently used for benchmarking convolutional neural networks~\cite{lecun2002gradient,xiao2017fashion,ciregan2012multi}.

Model Used: \textbf{VGG16}, a deep CNN known for its stacked convolutional layers~\cite{lecun2015deep}.

\paragraph{Semantic Segmentation Datasets}
\textbf{PASCAL VOC 2012}: A benchmark dataset for semantic segmentation, containing 20 object categories with pixel-wise annotations. Objects include people, animals, vehicles, and indoor objects, making it a challenging segmentation dataset~\cite{everingham2010pascal}.
Models Used: \textbf{FCN} (Fully Convolutional Network) + \textbf{ResNet50}: A model that replaces fully connected layers with convolutional layers for pixel-wise predictions~\cite{long2015fully}. DeepLabV3+: A state-of-the-art segmentation model using atrous convolutions and a spatial pyramid pooling module for fine-grained segmentation~\cite{chen2018encoder}.

\section{Logits Perturbation}
In Logits Perturbation, we add random Gaussian noise to the model's final output logits to affect its classification decisions. Specifically, we modify the original logits $z$ by adding noise drawn from a normal distribution with mean 0 and standard deviation $\sigma$.
\begin{align*}
    \tilde{z} = z+\mathcal{N}(0,\sigma^2)
\end{align*}
Here, $\sigma$ controls the strength of the noise. When $\sigma$ increases, the added noise causes more variation in the logits, making the SoftMax output more uncertain. As a result, the model's accuracy decreases. If $\sigma$ is small, the effect on logits is minimal, and the model can still classify correctly in most cases.
 
 \begin{figure}[h!]
    \centering
        \includegraphics[width=0.8\linewidth]{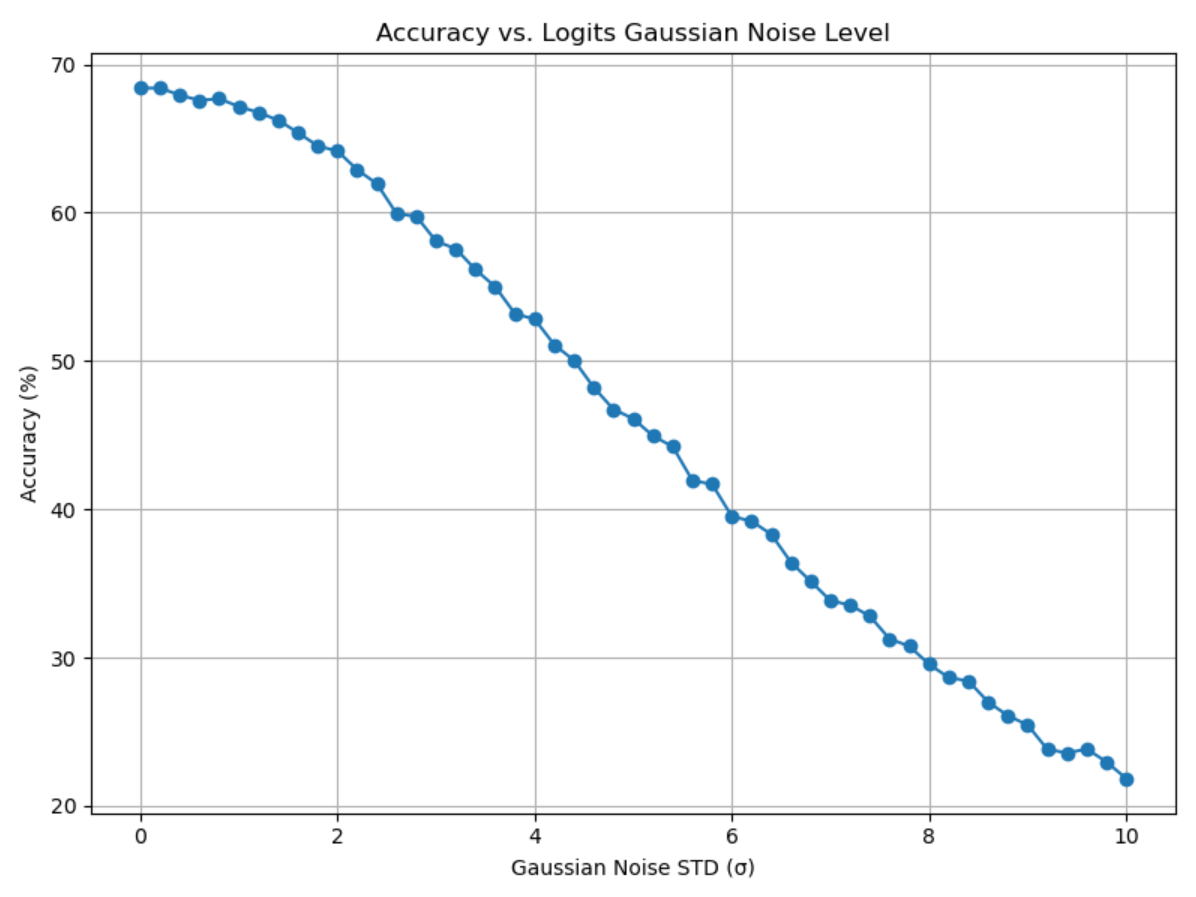}
    \caption{FCN ResNet50 on CIFAR100 dataset for Accuracy vs Noise level.}
    \label{fig:fig12}
\end{figure}

Fig.~\ref{fig:fig12} shows how classification accuracy changes under different levels of Gaussian noise added to the logits. The x-axis represents the standard deviation ($\sigma$) of the Gaussian noise, and the y-axis shows the corresponding accuracy in percentage. As the noise level increases, the accuracy gradually decreases, indicating a clear negative correlation. This trend demonstrates the model's sensitivity to noise.
  
 \begin{figure}[h!]
    \centering
        \includegraphics[width=0.8\linewidth]{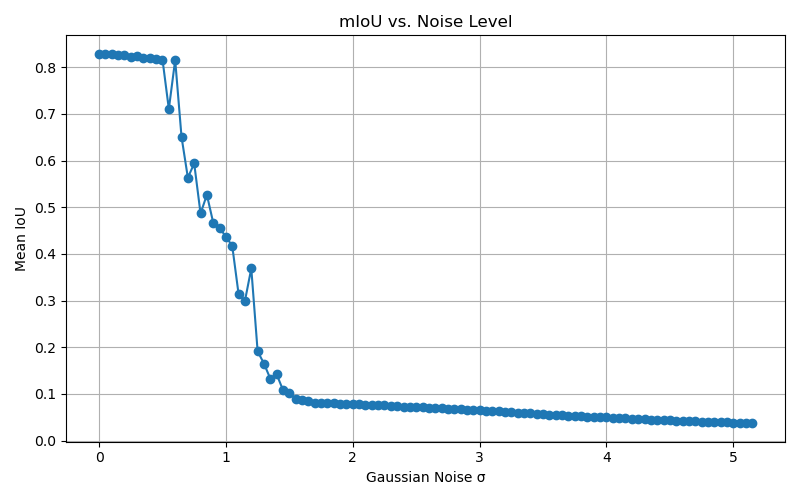}
    \caption{FCN ResNet50 on VOC 2012 dataset for Miou vs Noise level.}
    \label{fig:fig13}
\end{figure}

Fig.~\ref{fig:fig13} shows the effect of Gaussian noise on model performance in a semantic segmentation task. The x-axis represents the noise standard deviation ($\sigma$) added to the logits, and the y-axis shows the corresponding mean Intersection over Union (mIoU). At low noise levels, the model achieves high segmentation accuracy. However, once $\sigma$ exceeds around 1.0, the mIoU drops sharply, indicating a rapid loss in performance. It highlights how sensitive the segmentation model is to noise, and it helps assess its robustness under different noise conditions.
 
 \begin{figure}[h!]
    \centering
        \includegraphics[width=\linewidth]{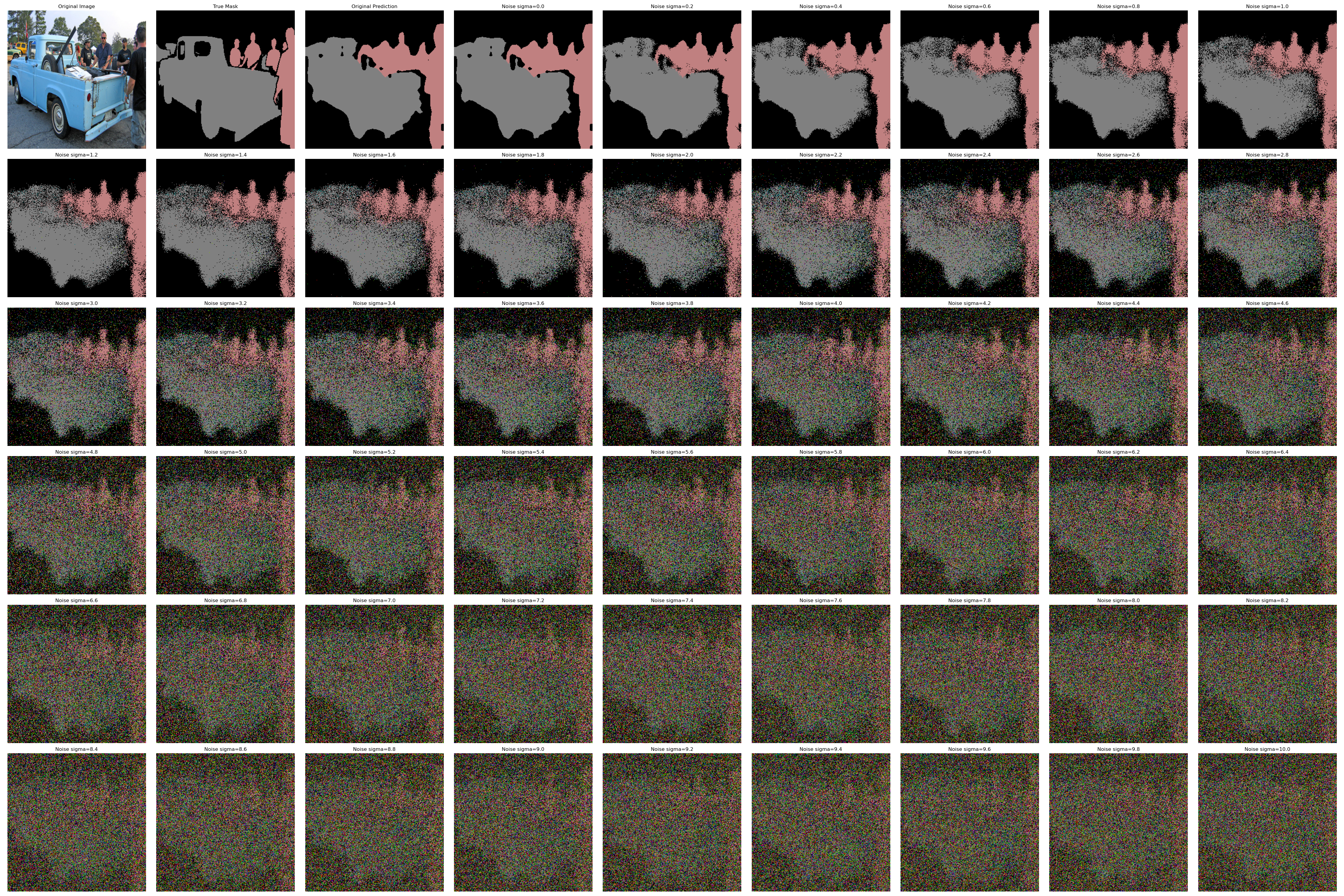}
    \caption{Semantic Segmentation task on FCN ResNet50 on VOC2012 with noise.}
    \label{fig:fig14}
\end{figure}

Fig.~\ref{fig:fig14} shows the impact of Gaussian noise on semantic segmentation results. The top-left image is the original input image, followed by the ground truth mask and the model's prediction without noise. The other columns show the results of segmentation after adding Gaussian noise with a higher standard deviation ($\sigma$) to the model's logits. The quality of the segmentation gets worse as the noise level goes from top to bottom. Even when there isn't much noise, the predicted masks still show clear edges between objects. But as $\sigma$ gets bigger, the masks get more and more messed up, which makes the segmentation blurry or wrong. This picture shows how Gaussian perturbation changes the performance of semantic segmentation models.

\section{Region-Focused Logits Perturbation}
The Region-Focused Logits Perturbation process begins by loading a pre-trained model, fine-tuned for semantic segmentation on the Pascal VOC dataset. The model takes an input image and predicts pixel-wise class probabilities, which are processed through a softmax function to generate an initial segmentation mask. We use Grad-CAM on the last convolutional layers to figure out which parts have the most significant effect on the model's decision. This makes a class activation map (CAM) that shows the most important parts of the image for the predicted segmentation. After that, a threshold is used to find the most important areas, which are then shown on a binary mask that shows where they are.
Adding Gaussian noise of different sizes ($\sigma$²) to the model's predictions in a way that isn't random comes after finding the most important parts. The noise is only added to the parts that Grad-CAM shows, so the less important parts stay the same. After that, the modified logits go through softmax again to make predictions about perturbed segmentation. We use the mean Intersection over Union (mIoU) and Dice coefficient metrics to compare the new predictions to the ground truth masks to see how these changes affect the results. The experiment shows how the model's robustness and the importance of certain areas in its decision-making process change when the noise level changes.
 
 \begin{figure}[h!]
    \centering
        \includegraphics[width=\linewidth]{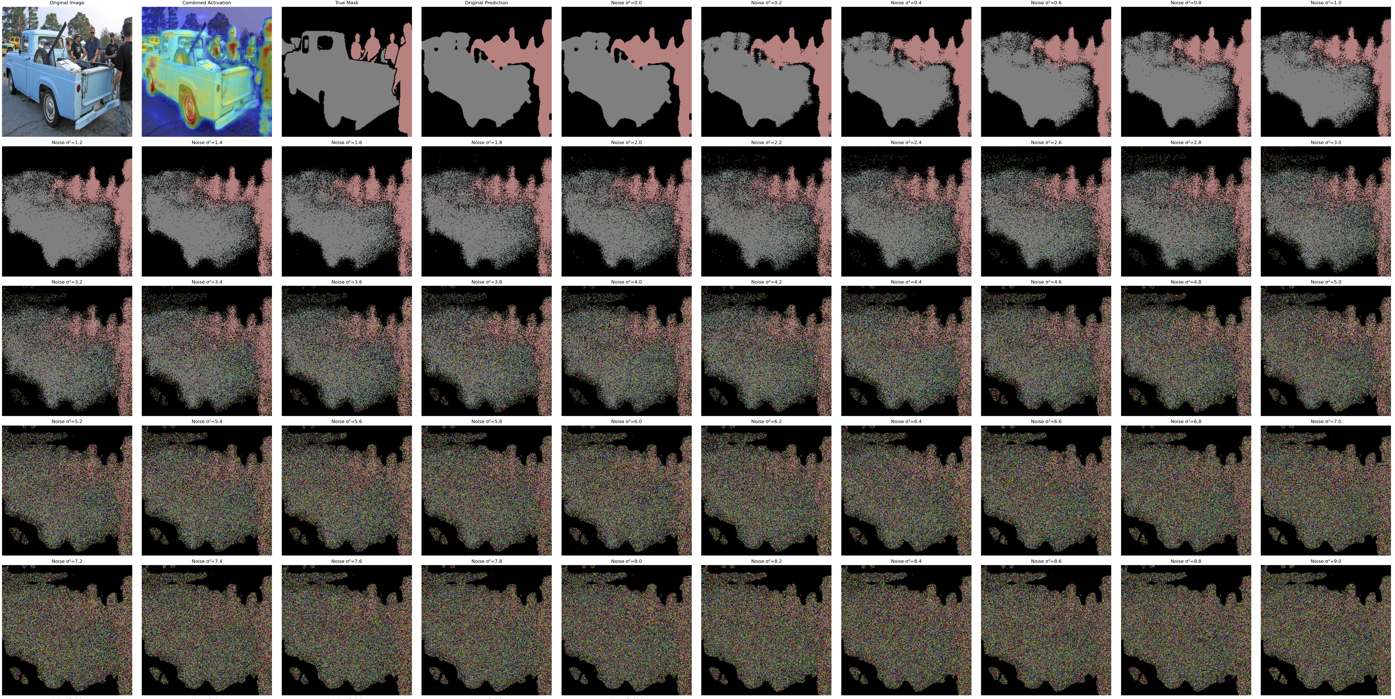}
    \caption{FCN ResNet50 using voc dataset with Grad.}
    \label{fig:fig15}
\end{figure}

Experiment demonstrates the impact of selectively adding noise to the most critical regions identified by Grad-CAM during semantic segmentation. The results (Fig.~\ref{fig:fig15}) clearly show that perturbing these regions leads to significant degradation in segmentation accuracy. As shown in Fig.~\ref{fig:fig15}, in the first row, an image of a truck is analyzed. The original image and its corresponding ground truth segmentation mask are shown alongside the Grad-CAM heatmap, which highlights the most influential regions for the model's prediction. The fourth column presents the model's initial segmentation output, which closely resembles the ground truth. However, after applying noise perturbation within the Grad-CAM-identified regions ($\sigma$² = 1.00), the segmentation mask becomes severely disrupted, with fragmented and misclassified areas appearing throughout the prediction.

 \begin{figure}[h!]
    \centering
        \includegraphics[width=\linewidth]{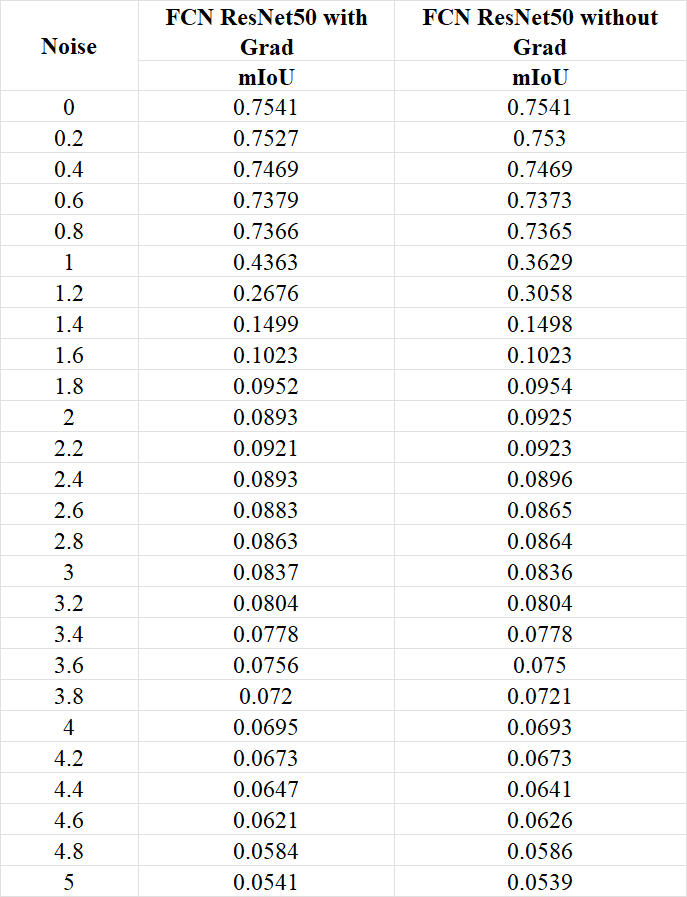}
    \caption{FCN ResNet50 using VOC dataset.}
    \label{fig:fig16}
\end{figure}

\begin{figure}[h!]
    \centering
        \includegraphics[width=\linewidth]{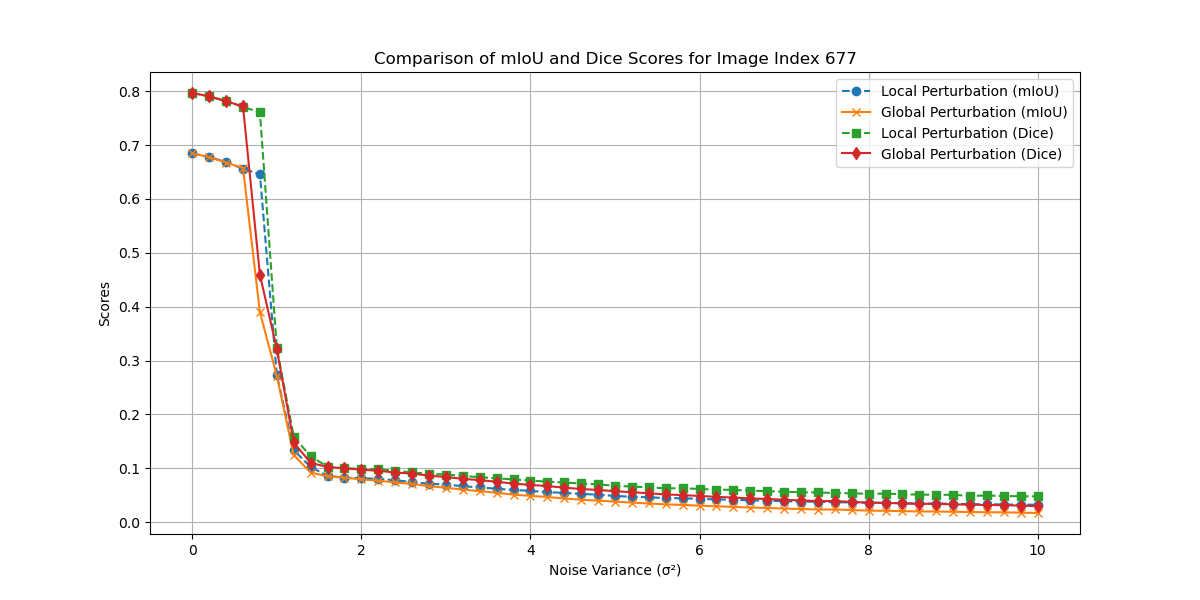}
    \caption{Comparison of MioU under Region-focused and Gloal Perturbation.}
    \label{fig:fig17}
\end{figure}

Table~\ref{fig:fig16} shows how the mIoU of FCN ResNet50 models changes under different levels of Gaussian noise ($\sigma$), with and without Grad-CAM guidance. As noise increases, mIoU gradually decreases for both models, indicating reduced segmentation accuracy. However, the model using Grad-CAM maintains slightly higher mIoU values across most noise levels, especially in the low-to-medium noise range. This suggests that Grad-CAM helps the model preserve performance under noise perturbation.
 Fig.~\ref{fig:fig17} Comparison of MioU under Region-focused and Gloal Perturbation
 
The comparison between Grad-CAM-guided perturbation and global perturbation highlights distinct differences in how noise affects the model's segmentation performance. Initially, both methods start with similar performance, achieving an mIoU of approximately 0.75 and a Dice coefficient around 0.85, indicating that the baseline segmentation predictions are accurate. However, as noise is introduced, their degradation patterns diverge.
In the Grad-CAM-based perturbation, where noise is selectively applied to the most critical regions identified by the model, the segmentation performance declines sharply even at low noise levels. By $\sigma$ = 0.80, the mIoU drops to 0.4363, while the globally perturbed model maintains a slightly better performance at 0.3629. This suggests that targeting high-importance regions has a more significant impact on the model's ability to make accurate predictions compared to distributing noise uniformly. As noise variance increases further, both methods experience a drastic reduction in segmentation quality. By $\sigma$ = 1.20, the mIoU in the Grad-CAM-guided approach falls below 0.15, indicating that the model struggles to maintain meaningful segmentation when its most informative regions are perturbed. The globally perturbed model exhibits a similar decline but retains marginally better performance for a short range before both methods converge.
At higher noise levels, specifically when $\sigma \geq 2.00$, the performance in both cases degrades to a point where segmentation becomes nearly indistinguishable, with mIoU dropping below 0.1 and Dice scores below 0.11. This indicates that when excessive noise is introduced, whether region-focused or global, the model loses its ability to produce reliable segmentation masks. However, the faster degradation observed in the Grad-CAM-guided perturbation reinforces the notion that the model relies heavily on a few key regions for decision-making. This suggests that the segmentation network is particularly vulnerable to adversarial attacks or perturbations in areas it deems important~\cite{liu2024bpkd}, emphasizing the need for improved robustness in deep learning-based semantic segmentation models.
 
\begin{figure}[h!]
    \centering
        \includegraphics[width=\linewidth]{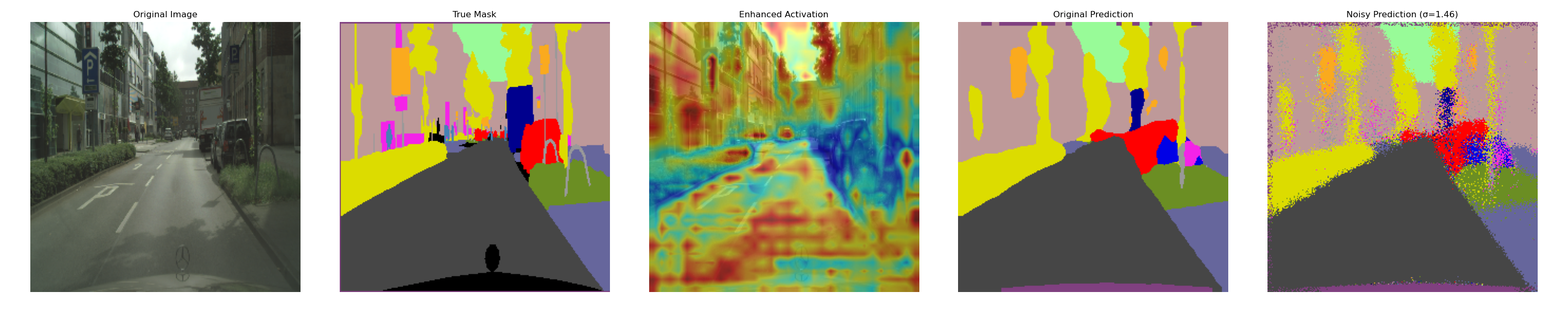}
    \caption{FCN ResNet50 using cityscape dataset with Grad.}
    \label{fig:fig18}
\end{figure}

Similarly, in the second example featuring an urban road scene (Fig.~\ref{fig:fig18}), the Grad-CAM activation map highlights key regions such as lane markings, vehicles, and buildings. The original prediction closely matches the ground truth, but once noise is introduced ($\sigma^2$ = 0.25), the segmentation output begins to deteriorate, particularly in the most crucial regions. Objects lose their defined boundaries, and smaller structures become indistinguishable. This further confirms that the segmentation model strongly depends on specific high-importance areas when making predictions.

Overall, the experiment reveals that region-focused perturbation is significantly more destructive than global noise since the most relevant features for segmentation are selectively disrupted. The rapid decline in segmentation performance demonstrates that the model is highly sensitive to perturbations in these areas, emphasizing the need for improved robustness mechanisms. These findings highlight the vulnerability of deep learning-based segmentation models to targeted adversarial noise and suggest that enhancing feature resilience in these key regions could be a crucial step toward developing more robust segmentation architectures.

The Region-Focused Logits Perturbation experiment shows how sensitive deep learning-based segmentation models are to noise, especially when it is applied to their most important regions. By adding noise specifically to areas highlighted by Grad-CAM, we can see that the model's predictions quickly become much worse. Even small amounts of noise cause objects in the segmentation masks to lose their shape, become scattered, or even disappear. This tells us that the model is strongly relying on a few key areas to make its decisions, and when those areas are disturbed, the entire prediction breaks down.

When we look at region-focused perturbation next to global perturbation, where noise is spread out over the whole image, we can see that targeting specific areas is much worse. When noise is only added to the important areas, the model's performance drops faster. This shows that these areas are very important for how the model understands the image. This is worrisome because in real life, like with self-driving cars, medical image analysis, and satellite imaging, small changes in important areas could cause big problems. It's easy to trick or confuse a model if it relies heavily on just a few parts of an image. This is a major flaw.

You can learn from this experiment that Grad-CAM and other explainability tools can help you find these flaws. We can use this information to make models that work better because they tell us which parts of an image are most important to the model. Researchers should find ways to make models less reliant on small areas and better at handling noise in the future. One way to do this is to use training methods that add noise while learning. Another is to make models better at focusing on different parts of an image. A third way is to design architectures that spread out important features instead of relying on a few specific areas.
In short, Region-Focused Logits Perturbation shows that segmentation models have a big problem: they rely too much on certain key regions, which makes them easy to break. If we want these models to be more reliable in real-world applications, we need to work on making them more stable and less sensitive to small changes in their most important features.

\section{Mapping via a Fitted Function}

In the experiment, we use an optimization method to find the right $\sigma$ value that reduces accuracy to a target level. For example, if we want to decrease accuracy from 90\% to 50\%, we search for the noise level that achieves this effect. Logits Perturbation allows us to precisely control the model's performance, adjusting accuracy based on different needs.
 
\begin{figure}[h!]
    \centering
        \includegraphics[width=\linewidth]{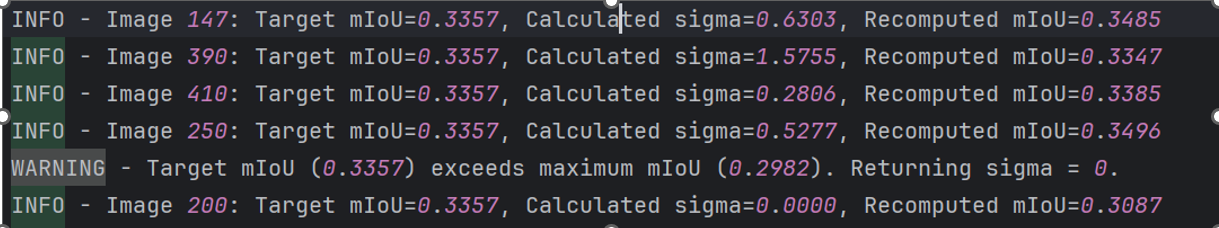}
    \caption{Target mIoU vs. achieved mIoU for FCN-ResNet50 on the PASCAL VOC dataset.}
    \label{fig:fig19}
\end{figure}

Fig.~\ref{fig:fig19} shows the comparison between target mIoU and the actual mIoU after applying noise with a calculated sigma value using FCN-ResNet50 on the PASCAL VOC dataset. For each image, the system estimates the noise level ($\sigma$) needed to reach the target mIoU (0.3357). Most recomputed mIoU values are close to the target, showing the fitted function works well. In one case, the target mIoU was higher than the model's maximum possible output, so sigma was set to zero and the original prediction was used.
 
\begin{figure}[h!]
    \centering
        \includegraphics[width=\linewidth]{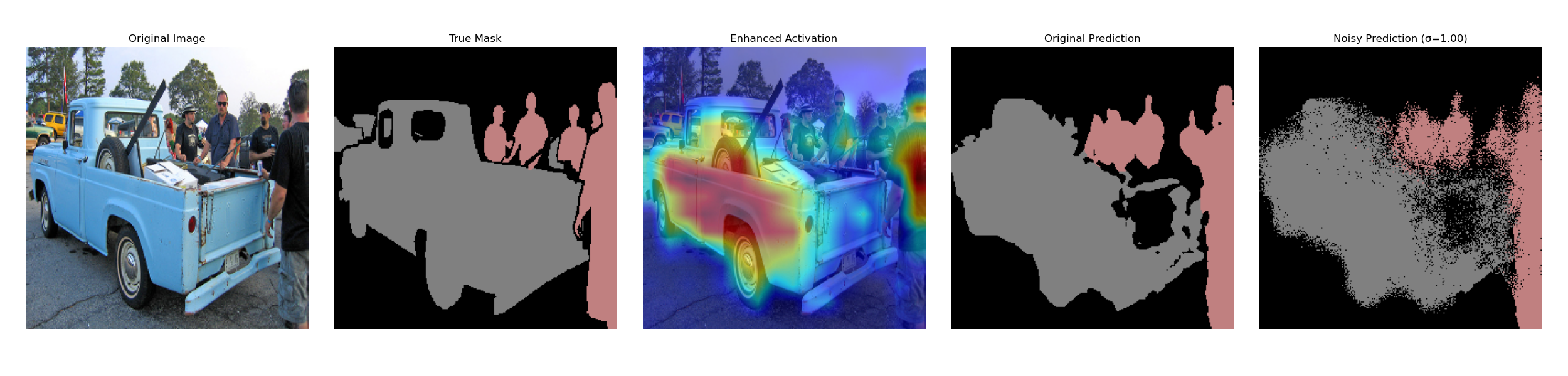}
    \caption{FCN ResNet50 using voc dataset with Grad, given target MioU.}
    \label{fig:fig20}
\end{figure}

Fig.~\ref{fig:fig20} shows the process of region-focused perturbation applied to semantic segmentation using FCN ResNet50 on the VOC dataset, given a target Miou. From left to right, we see the original image, the ground truth mask, and the enhanced activation map generated by Grad-CAM. The fourth image shows the original segmentation prediction, while the last image displays the noisy prediction under a set noise level ($\sigma$ = 1.00). Compared to the original output, the noisy prediction becomes more blurred, especially in areas with high importance, indicating that perturbation is effectively guided by the attention map.

\chapter{Conclusion}
The \textsc{NNObfuscator} framework is designed to meet two key requirements: Personalized Adaptation for Users and Strategic Model Tiering for Owners, while ensuring that the underlying model structure remains unchanged. Unlike methods such as fine-tuning, transfer learning, or feature distillation, which modify the internal weights or architecture of a model, \textsc{NNObfuscator} works by selectively obfuscating critical features in the model’s learned representations. This allows it to adapt model outputs based on user-specific needs.

Our tests show that selectively hiding important features, like those found with Grad CAM, has a big effect on how well the model predicts. By using region-focused perturbation, we can carefully lower a model's accuracy in a controlled way, either by customizing outputs for different users or by tiering model performance based on access levels. This is not the same as global perturbation methods, which make performance worse in ways that are hard to predict. The results show that deep learning models often depend a lot on a small number of very important features. \textsc{NNObfuscator} can change these features in a targeted way to control how the model behaves without having to retrain or change the base model.

However, it can be hard to make obfuscation work well and have a purpose while still keeping the model usable. One of the biggest problems is making sure that feature obfuscation is set up correctly for different applications, whether for personalized user experiences or strategic access control. Also, different models and tasks may be more or less sensitive to obfuscation, which means that adaptive techniques may be needed to get the best performance in different situations.

\textsc{NNObfuscator} gives you a new way to control and change deep learning models without changing how they work inside. It lets users have personalized experiences while giving model owners strategic control over performance tiers by using targeted obfuscation. This method is flexible and efficient for changing AI models to meet different needs while still keeping them safe, fair, and easy to use.

\chapter{Future Work}
To further explore the generalizability of our method beyond classification and segmentation tasks, we conducted a similar experiment on a generative model Stable Diffusion 1.5~\cite{ho2020denoising}, a popular text-to-image diffusion model~\cite{rombach2022high,saharia2022photorealistic}. In this experiment, we used the prompt ``two dogs running on the grass'' to generate a baseline image and then introduced Gaussian noise at the latent stage just before decoding. The goal was to observe how different levels of latent space perturbation would affect the visual quality of the generated images.
 
\begin{figure}[h!]
    \centering
        \includegraphics[width=\linewidth]{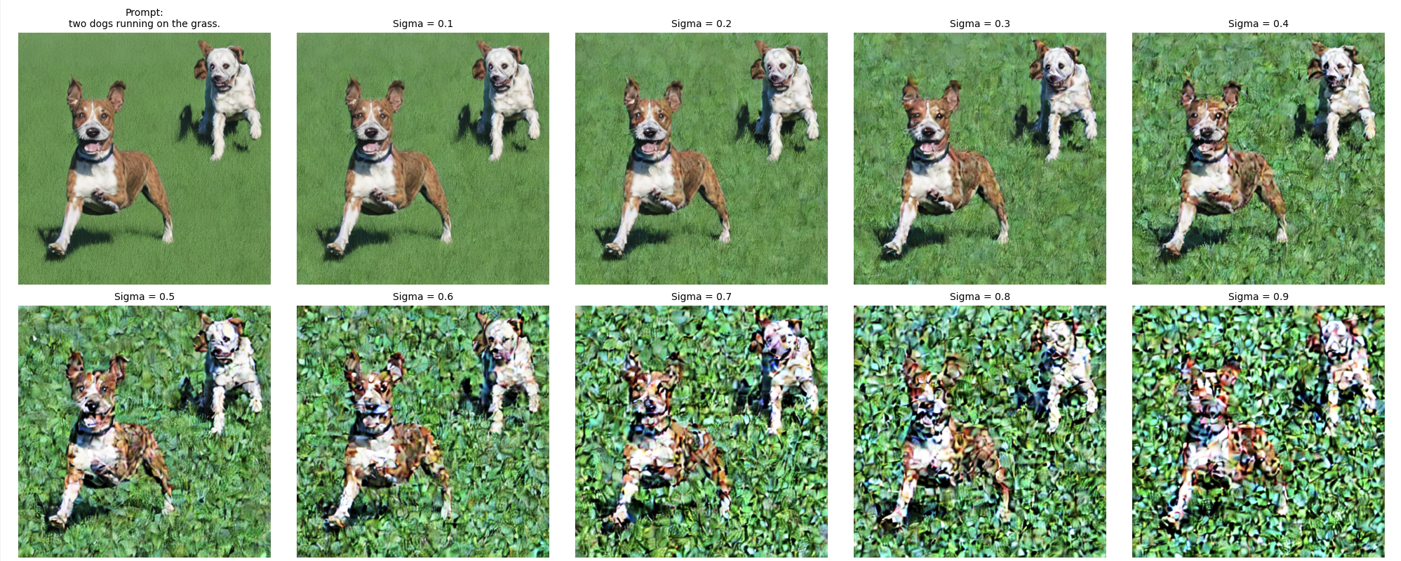}
    \caption{Stable Diffusion with Gaussian Noise.}
    \label{fig:fig21}
\end{figure}

We tested a series of standard deviation values ranging from $\sigma$ = 0.1 to $\sigma$ = 0.9. The results are shown in the figure above (Fig.~\ref{fig:fig21}). When the noise level was low (e.g., $\sigma \leq$ 0.3), the images generated remained visually consistent and recognizable, with minimal distortion. As the noise increased ($\sigma$ between 0.4 and 0.6), noticeable artifacts began to appear, especially in fine details and texture consistency. At higher levels ($\sigma \geq$ 0.7), the images became heavily degraded, with shapes and objects losing coherence, while the overall structure collapsed into noise-like patterns. These visual changes clearly demonstrate a progressive loss in generation fidelity as noise increases.
This experiment shows that latent space perturbation can be an effective way to adjust the output quality of generative models in a controlled manner. It confirms that our proposed utility control method is not limited to tasks such as classification and segmentation, but also applies to generative tasks. Moreover, the same methodology can be extended to other types of generative models, such as text-to-speech systems, image inpainting models, and diffusion-based video generation frameworks. By injecting controlled noise into the latent representations, one can modulate the output quality in a predictable way, which is useful for applications that require tiered content generation, robustness testing, or privacy-preserving outputs.

\textbf{Limitation}. While our approach enables flexible utility control and supports tiered service levels for generative models, it also introduces new security and privacy risks that warrant careful consideration. Specifically, exposing multiple utility tiers may increase vulnerability to model extraction or reconstruction attacks, where adversaries aggregate outputs from lower-tier models to approximate the performance of the full-capability model without proper authorization. This risk is especially relevant in generative tasks, where subtle differences in output quality might be exploited by adaptive attackers through techniques such as collusion, model inversion, or membership inference~\cite{shokri2017membership,hu2022m,liu2025unlock,feng2024grab,ma2023loden}. Additionally, injecting controlled noise into the latent space could potentially leak information about the original, high-utility model if not carefully managed.

Addressing these risks requires further research into secure obfuscation techniques and robust access control mechanisms~\cite{wang2024core}, ensuring that utility modulation does not inadvertently compromise model confidentiality or user privacy. Future work should investigate defense strategies such as differential privacy, output randomization, purpose limitation, or secure model APIs to mitigate potential abuse in multi-tiered AI services.

\chapter*{Acknowledgment}
I would like to thank my supervisor A/Prof. Guangdong Bai, and my mentor, Zihan Wang, for their valuable guidance and continuous support throughout this research. Their helpful advice, support, and patience helped me stay on track and make my work better bit by bit. I also appreciate how my family and friends have always been there for me, which helped me stay focused on this project.


\bibliographystyle{IEEEtran}
\bibliography{references}

\end{document}